\documentclass[11pt,a4paper]{article}
\usepackage[hyperref]{acl2017}
\usepackage{times}
\usepackage{latexsym}
\aclfinalcopy
\usepackage{url}

\usepackage{amsmath}
\usepackage{amssymb}
\DeclareMathOperator*{\argmax}{argmax}
\DeclareMathOperator*{\Softmax}{Softmax}

\usepackage{multirow}

\usepackage{times}
\usepackage{graphics}
\usepackage{epsfig}
\usepackage{amsfonts}
\usepackage{graphicx}
\usepackage{makecell}
\usepackage{indentfirst}

\usepackage[ruled,vlined,linesnumbered]{algorithm2e}

\title{Jointly Extracting Relations with Class Ties\\via Effective Deep Ranking}

\author{Hai Ye$^1$, Wenhan Chao$^1$, Zhunchen Luo$^2$\thanks{ \ \ \ Corresponding author. Codes are available at \url{https://github.com/oceanypt/DR_RE}.},\ Zhoujun Li$^1$\\ 
$^1$School of Computer Science and Engineering, Beihang University, Beijing 100191, China\\
\{yehai, chaowenhan, lizj\}@buaa.edu.cn \\
$^2$China Defense Science and Technology Information Center, Beijing 100142, China\\
zhunchenluo@gmail.com
}

\date{}

\begin{document}
\maketitle
\begin{abstract}
Connections between relations in relation extraction, which we call \emph{class ties}, are common. In distantly supervised  scenario, one entity tuple may have multiple relation facts. 
Exploiting class ties between relations of one entity tuple will be promising for distantly supervised relation extraction. However, previous models are not effective or ignore to model this property. In this work, to effectively leverage class ties, we propose to make joint relation extraction with a unified model that integrates convolutional neural network (CNN) with a general pairwise ranking framework, in which three novel ranking loss functions are introduced. Additionally, an effective method is presented to relieve the severe class imbalance problem from NR (not relation) for model training. 
Experiments on a widely used dataset show that leveraging class ties will enhance extraction and demonstrate the effectiveness of our model to learn class ties. Our model outperforms the baselines significantly, achieving state-of-the-art performance.
\end{abstract}

\section{Introduction}
Relation extraction (RE) 
aims to classify the relations between two given named entities from natural-language text. 
Supervised machine learning methods require numerous labeled data to work well. 
With the rapid growth of volume of relation types, traditional methods can not keep up with the step for the limitation of labeled data. In order to narrow down the gap of data sparsity, \citet{2009} propose \emph{distant supervision (DS)} for relation extraction, which automatically generates training data by aligning a knowledge facts database (ie. Freebase \cite{freebase}) with texts. 

\emph{Class ties} mean the connections between relations in relation extraction. 
In general, we conclude that class ties can have two types: weak class ties and strong class ties. Weak class ties mainly involve the co-occurrence of relations such as \emph{place\_of\_birth} and \emph{place\_lived}, \emph{CEO\_of} and \emph{founder\_of}. On the contrary, strong class ties mean that relations have latent logical entailments. Take the two relations of \emph{capital\_of} and \emph{city\_of} for example, if one entity tuple has the relation of \emph{capital\_of}, it must express the relation fact of \emph{city\_of}, because the two relations have the entailment of \emph{capital\_of} $\Rightarrow$ \emph{city\_of}. Obviously the opposite induction is not correct.
Further take the sentence of \emph{``Jonbenet told me that her mother $\text{[Patsy Ramsey]}_{e_1}$ never left $\text{[Atlanta]}_{e_2}$ since she was born.''} in DS scenario for example.
This sentence expresses two relation facts which are \emph{place\_of\_birth} and \emph{place\_lived}.
However, the word ``born" is a strong bios to extract \emph{place\_of\_birth}, so it may not be easy to predict the relation of \emph{place\_lived}, but if we can incorporate the weak ties between the two relations, extracting \emph{place\_of\_birth} will provide evidence for prediction of \emph{place\_lived}.

\begin{table}
\small
  \centering
  \begin{tabular}{c}
 \emph{place\_lived (Patsy Ramsey, Atlanta)}   \\ 
 \emph{place\_of\_birth (Patsy Ramsey, Atlanta)} \\
  \end{tabular}
  \begin{tabular}{|c|p{4.cm}|p{1.65cm}|}
  \hline
   & Sentence & Latent Label \\ \hline
   $\#1$ & \emph{Patsy Ramsey} has been living in \emph{Atlanta} since she was born. & \emph{place\_of\_birth} \\ \hline
   $\#2$ & \emph{Patsy Ramsy} always loves \emph{Atlanta} since it is her hometown. & \emph{place\_lived} \\ \hline
  \end{tabular}
  \caption{Training instances generated by freebase.}\label{Tab1}
\end{table}


Exploiting class ties is necessary for DS based relation extraction. In DS scenario, there is a challenge that one entity tuple can have multiple relation facts as shown in Table \ref{Tab1}, which is called \emph{relation overlapping} \cite{2011,2012}. However, the relations of one entity tuple can have class ties mentioned above which can be leveraged to enhance relation extraction for it narrowing down potential searching spaces and reducing uncertainties between relations when predicting unknown relations. If one pair entities has \emph{CEO\_of}, it will contain \emph{founder\_of} with high possibility. 

To exploit class ties between relations, we propose to make joint extraction for all positive labels of one entity tuple with considering \emph{pairwise} connections between positive and negative labels inspired by \cite{furnkranz2008multilabel,zhang2006multilabel}. As the two relations with class ties shown in Table \ref{Tab1}, by joint extraction of two relations, we can maintain the \emph{class ties} (co-occurrence) of them from training samples to be learned by potential model, and then leverage this learned information to extract instances with unknown relations, which can not be achieved by separated extraction for it dividing labels apart losing information of co-occurrence. 
To classify positive labels from negative ones, we adopt pairwise ranking to rank positive ones higher than negative ones, exploiting pairwise connections between them. 
In a word, joint extraction exploits class ties between relations and pairwise ranking classify positive labels from negative ones.
Furthermore, combining information across sentences will be more appropriate for joint extraction which provides more information from other sentences to extract each relation \cite{hao,lin2016}. In Table \ref{Tab1}, sentence \#1 is the evidence for \emph{place\_of\_birth}, but it also expresses the meaning of ``living in someplace'', so it can be aggregated with sentence \#2 to extract \emph{place\_lived}. Meanwhile, the word of ``hometown'' in sentence \#2 can provide evidence for \emph{place\_of\_birth} which should be combined with sentence \#1 to extract \emph{place\_of\_birth}.

\begin{figure}
\centering\includegraphics[width = 6.3cm]{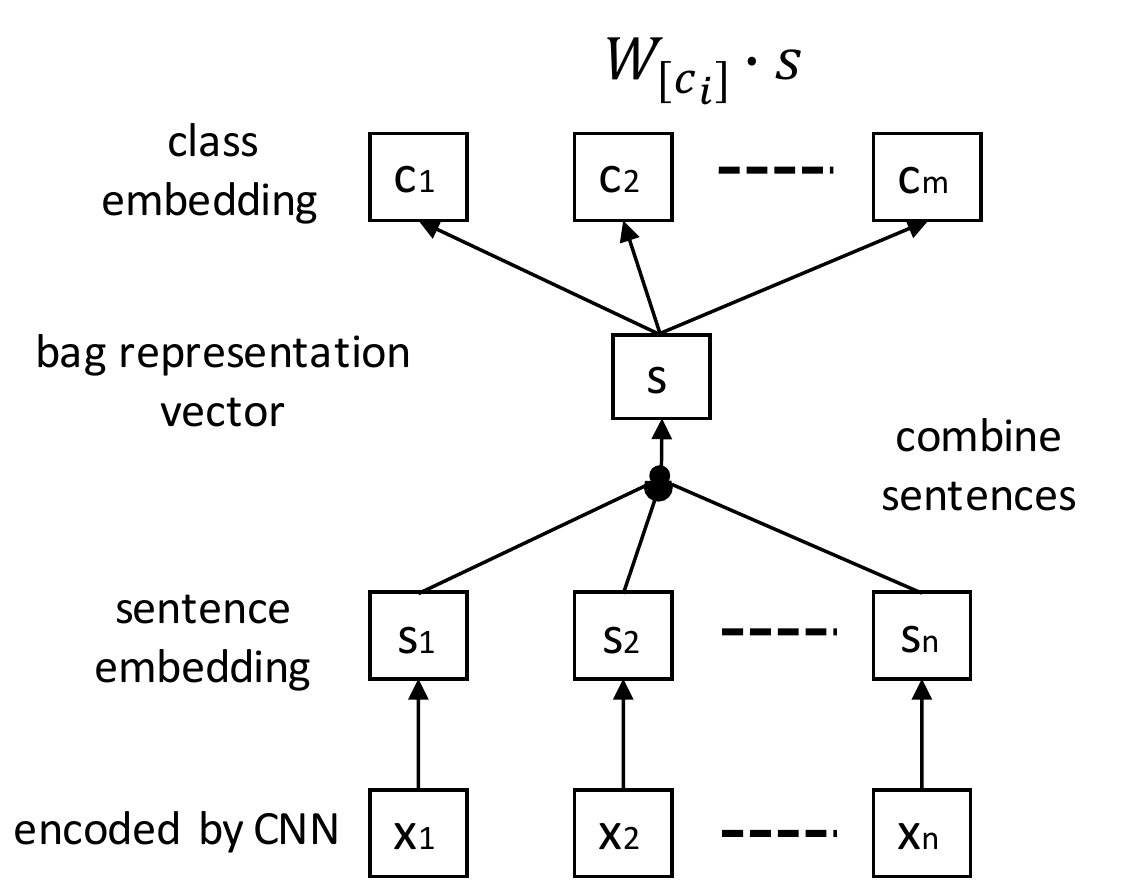}
\caption{The main architecture of our model.}
\label{archi}
\end{figure}

In this work, we propose a unified model that integrates pairwise ranking with CNN to exploit class ties. 
Inspired by the effectiveness of deep learning for modeling sentences \cite{deeplearning2015}, we use CNN to encode sentences. Similar to \cite{rank2015, lin2016}, we use class embeddings to represent relation classes. 
The whole model architecture is presented in Figure \ref{archi}. 
We first use CNN to embed sentences, 
then we introduce two variant methods to combine the embedded sentences into one bag representation vector aiming to aggregate information across sentences, 
after that we measure the similarity between bag representation and relation class in real-valued space.
With two variants for combining sentences, three novel pairwise ranking loss functions are proposed to make joint extraction.
Besides, to relieve the bad impact of class imbalance from NR (not relation) \cite{japkowicz2002class} for training our model, we cut down loss propagation from NR class during training.  

Our experimental results on dataset of \citet{2010} are evident that: (1) Our model is much more effective than the baselines; (2) Leveraging class ties will enhance relation extraction and our model is efficient to learn class ties by joint extraction; (3) A much better model can be trained after relieving class imbalance from NR.

Our contributions in this paper can be encapsulated as follows:

$\bullet$ We propose to leverage class ties to enhance relation extraction. An effective deep ranking model which integrates CNN and pairwise ranking framework is introduced to exploit class ties. 

$\bullet$ We propose an effective method to relieve the impact of data imbalance from NR for model training. 

$\bullet$ Our method achieves state-of-the-art performance.

\section{Related Work}
We summarize related works on two main aspects: 
\subsection{Distant Supervision Relation Extraction}
Previous works on DS based RE ignore or are not effective to leverage class ties between relations. 

 \citet{2010} introduce multi-instance learning to relieve the wrong labelling problem, ignoring class ties.  Afterwards, \citet{2011} and \citet{2012} model this problem by multi-instance multi-label learning to extract overlapping relations. Though they also propose to make joint extraction of relations, they only use information from single sentence losing information from other sentences. \citet{Global} try to use \emph{Markov logic} model to capture consistency between relation labels, on the contrary, our model leverages deep ranking to learn class ties automatically.

With the remarkable success of deep learning in CV and NLP \cite{deeplearning2015}, deep learning has been applied to relation extraction \cite{zeng2014,zeng2015,rank2015,lin2016}, 
the specific deep learning architecture can be CNN \cite{zeng2014}, RNN \cite{zhou2016attention}, etc. \citet{zeng2015} propose a piecewise convolutional neural network with multi-instance learning for DS based relation extraction, which improves the precision and recall significantly. Afterwards, \citet{lin2016} introduce the mechanism of attention \cite{attention1,attention2} to select the sentences to relieve the wrong labelling problem and use all the information across sentences. However, the two deep learning based models only make separated extraction thus can not model class ties between relations.

\subsection{Deep Learning to Rank}
Deep learning to rank has been widely used in many problems to serve as a classification model. In image retrieval, \citet{zhao2015deep} apply deep semantic ranking for multi-label image retrieval. In text matching, \citet{severyn2015learning} adopt learning to rank combined with deep CNN for short text pairs matching. In traditional supervised relation extraction, \citet{rank2015} design a pairwise loss function based on CNN for single label relation extraction. Based on the advantage of deep learning to rank, we propose pairwise learning to rank (LTR) \cite{liu2009} combined with CNN in our model aiming to jointly extract multiple relations. 



\section{Proposed Model}
In this section, we first conclude the notations used in this paper, then we introduce the used CNN for sentence embedding, afterwards, we present our algorithm of how to learn class ties between relations of one entity tuple. 

\subsection{Notation}
We define the relation classes as $\mathcal{L} = \{1,2,\cdots,C\}$, entity tuples as $\mathcal{T}=\{t_i\}_{i=1}^M$ and mentions\footnote{The sentence containing one certain entity is called mention.} as $\mathcal{X} = \{x_i\}_{i=1}^N$. Dataset is constructed as follows: for entity tuple $t_i \in \mathcal{T}$ and its relation class set $L_i \subseteq \mathcal{L}$, we collect all the mentions $X_i$ that contain $t_i$, the dataset we use is $\mathcal{D} = \{(t_i, L_i, X_i)\}_{i=1}^H$.
Given a data $(t_k, L_k, X_k) \in \{(t_i, L_i, X_i)\}_{i=1}^H$, the sentence embeddings of $X_k$ encoded by CNN are defined as $S_k = \{{s_i}\}_{i=1}^{|X_k|}$ and 
we use class embeddings $W \in R^{|\mathcal{L}| \times d}$ to represent the relation classes.


\subsection{CNN for Sentence Embedding}
We take the effective CNN architecture adopted from \cite{zeng2015,lin2016} to encode sentence and we briefly introduce CNN in this section. More details of our CNN can be obtained from previous work.

\subsubsection{Words Representations}

\noindent{$\bullet$ \textbf{Word Embedding}  Given a word embedding matrix $V \in \mathbb{R}^{l^w \times d^1}$ where $l^w$ is the size of word dictionary and $d^1$ is the dimension of word embedding, the words of a mention $x = \{w_1, w_2, \cdots, w_n\}$ will be represented by real-valued vectors from $V$.}

\noindent{$\bullet$ \textbf{Position Embedding} The position embedding of a word measures the distance from the word to entities in a mention.
We add position embeddings into words representations by appending position embedding to word embedding for every word. Given a position embedding matrix $P \in \mathbb{R}^{l^p \times d^2}$ where $l^p$ is the number of distances and $d^2$ is the dimension of position embeddings, the dimension of words representations becomes $d^w = d^1 + d^2 \times 2$.}

\subsubsection{Convolution, Piecewise max-pooling}

After transforming words in $x$ to real-valued vectors, we get the sentence $q \in {\mathbb{R}}^{n \times d^w}$. The set of kernels $K$ is $\{ {K_i}\}_{i=1}^{d^s}$ where $d^s$ is the number of kernels. Define the window size as $d^{win}$ and given one kernel $K_k \in {\mathbb{R}}^{d^{win} \times d^w}$, the convolution operation is defined as follows:
\begin{equation}
m_{[i]} = q_{[i:i+d^{win}-1]} \odot K_k + b_{[k]}
\end{equation}
where $m$ is the vector after conducting convolution along $q$ for $n - d^{win} + 1$ times and $b \in {\mathbb{R}}^{d^s}$ is the bias vector.
For these vectors whose indexes out of range of $[1,n]$, we replace them with zero vectors.



By piecewise max-pooling, when pooling, the sentence is divided into three parts: $m_{[p_0:p_1]}$, $m_{[p_1:p_2]}$ and $m_{[p_2:p_3]}$ ($p_1$ and $p_2$ are the positions of entities, $p_0$ is the beginning of sentence and $p_3$ is the end of sentence). This piecewise max-pooling is defined as follows:
\begin{equation}
z_{[j]} = max(m_{[p_{j-1}:p_j]})
\end{equation}
where $z \in \mathbb{R}^3$ is the result of mention $x$ processed by kernel $K_k$; $1 \le j \le 3$. 
Given the set of kernels ${K}$, following the above steps, the mention $x$ can be embedded to $o$ where $o \in R^{d^s * 3}$.
\subsubsection{Non-Linear Layer, Regularization}
To learn high-level features of mentions, we apply a non-linear layer after pooling layer. After that, a dropout layer is applied to prevent over-fitting. We define the final fixed sentence representation as $s \in {\mathbb{R}}^{d^f}$ ($d^f = d^s * 3$).
\begin{equation}
s = g(o) \circ h
\end{equation}
where $g(\cdot)$ is a non-linear function and we use $tanh(\cdot)$ in this paper; $h$ is a Bernoulli random vector with probability p to be $1$.

\subsection{Learning Class Ties by Joint Extraction with Pairwise Ranking}
\begin{figure}
\centering\includegraphics[width = 6.5cm]{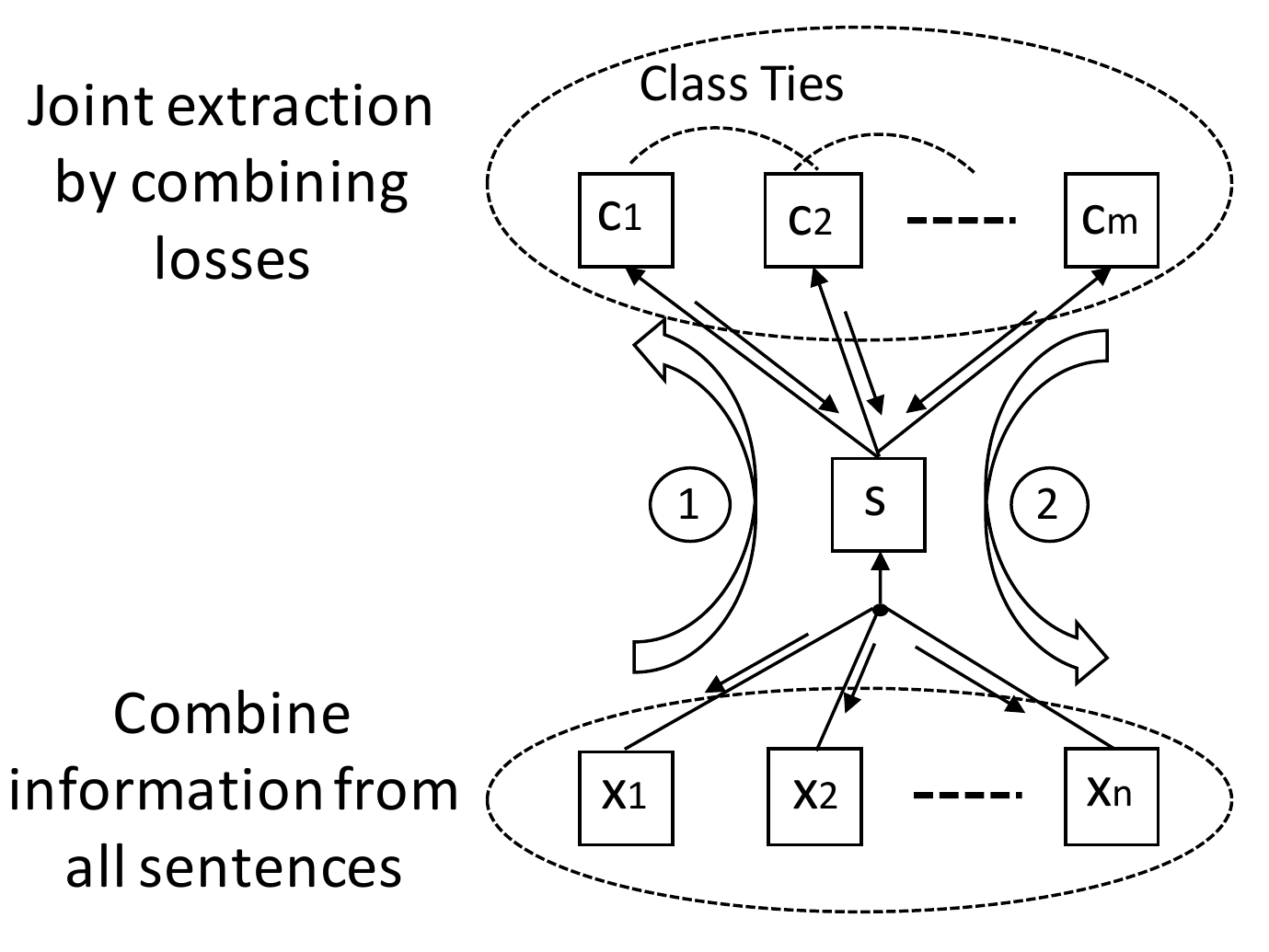}
\caption{Illustration of mechanism of our model to model class ties between relations.}
\label{archi2}
\end{figure}

As mentioned above, to learn class ties, we propose to make joint extraction with considering pairwise connections between positive labels and negative ones. Pairwise ranking is applied to achieve this goal. Besides, combining information across sentences is necessary for joint extraction. More specifically, as shown in Figure \ref{archi2}, from down to top, all information from sentences is pre-propagated to provide enough information for joint extraction. From top to down, pairwise ranking jointly extracting positive relations by combining losses, which are back-propagated to CNN to learn class ties. 

\subsubsection{Combining Information across Sentences}
We propose two options to combine sentences to provide enough information for joint extraction.

\noindent{$\bullet$ \textbf{AVE} \ \ The first option is average method. This method regards all the sentences equally and directly average the values in all dimensions of sentence embedding. This \textbf{AVE} function is defined as follows:}
\begin{equation}
\label{func4}
s = \frac{1}{n}\sum_{s_i \in S_k} s_i
\end{equation}
where $n$ is the number of sentences and $s$ is the representation vector combining all sentence embeddings.
Because it weights the importance of sentences equally, this method may bring much noise data from two aspects: (1) the wrong labelling data; (2) irrelated mentions for one relation class, for all sentences containing the same entity tuple being combined together to construct the bag representation. 

\noindent{$\bullet$ \textbf{ATT} \ \ The second one is a sentence-level attention algorithm used by \citet{lin2016} to measure the importance of sentences aiming to relieve the wrong labelling problem. For every sentence, \textbf{ATT} will calculate a weight by comparing the sentence to one relation. We first calculate the similarity between one sentence embedding and relation class as follows:}
\begin{equation}
e_j = a \cdot W_{[c]} \cdot s_j
\label{func5}
\end{equation}
where $e_j$ is the similarity between sentence embedding $s_j$ and relation class $c$ and a is a bias factor. In this paper, we set $a$ as $0.5$. Then we apply $\Softmax$ to rescale $\mathbf{e}$ ($\mathbf{e} = \{e_i\}_{i=1}^{|X_k|}$) to $[0,1]$. We get the weight ${\alpha}_j$ for $s_j$ as follows:
\begin{equation}
{\alpha}_j = \frac{\exp(e_j)}{\sum_{e_i \in \mathbf{e}} \exp(e_i)}
\label{func6}
\end{equation}
so the function to merge $s$ with \textbf{ATT} is as follows:
\begin{equation}
s = \sum_{i=1}^{|X_k|} \alpha_i \cdot s_i
\label{func7}
\end{equation}


\subsubsection{Joint Extraction by Combining Losses to Learn Class Ties}
Firstly, we have to present the score function to measure the similarity between $s$ and relation $c$.

\noindent{$\bullet$ \textbf{Score Function} \ \ We use dot function to produce score for $s$ to be predicted as relation $c$. The score function is as follows:}
\begin{equation}
\mathcal{F}(s,c) = W_{[c]}\cdot s
\end{equation}

There are other options for score function. In \citet{MultiATT}, they propose a margin based loss function that measures the similarity between $s$ and $W_{[c]}$ by distance. Because score function is not an important issue in our model, we adopt dot function, also used by \citet{rank2015} and \citet{lin2016}, as our score function.

Now we start to introduce the ranking loss function.

Pairwise ranking aims to learn the score function $\mathcal{F}(s,c)$ that ranks positive classes higher than negative ones. This goal can be summarized as follows:
\begin{equation}
 \forall c^+ \in L_k, \forall c^- \in \mathcal{L}-L_k : {\mathcal{F}}(s,c^+) > {\mathcal{F}}(s,c^-) + \beta
 \label{frank}
\end{equation}
where $\beta$ is a margin factor which controls the minimum margin between the positive scores and negative scores.

To learn class ties between relations, we extend the formula (\ref{frank}) to make joint extraction and we propose three ranking loss functions with variants of combining sentences. Followings are the proposed loss functions: 


\noindent{$\bullet$ \textbf{with AVE (Variant-1)} \ \ We define the margin-based loss function with option of \textbf{AVE} to aggregate sentences as follows:}
\begin{gather}
\nonumber G_{[ave]} = \sum_{c^+ \in L_k} \rho[0, \sigma^+ - \mathcal{F}(s,c^+)]_+ \\+ \rho|L_k|[0, \sigma^- + \mathcal{F}(s,c^-)]_{+}
\label{func9}
\end{gather}
where $[0, \cdot \ ]_+ = max(0,\cdot \ )$; $\rho$ is the rescale factor, $\sigma^+$ is positive margin and $\sigma^-$ is negative margin.
Similar to \citet{rank2015} and \citet{MultiATT}, this loss function is designed to rank positive classes higher than negative ones controlled by the margin of $\sigma^+ - \sigma^-$.
In reality, $\mathcal{F}(s,c^+)$ will be higher than $\sigma^+$ and $\mathcal{F}(s,c^-)$ will be lower than $\sigma^-$. In our work, we set $\rho$ as $2$, $\sigma^+$ as $2.5$ and $\sigma^-$ as $0.5$ adopted from \citet{rank2015}.

Similar to \citet{Weston2011} and \citet{rank2015}, we update one negative class at every training round but to balance the loss between positive classes and negative ones, we multiply $|L_k|$ before the right term in function (\ref{func9}) to expand the negative loss. We apply mini-bach based stochastic gradient descent (SGD) to minimize the loss function. The negative class is chosen as the one with highest score among all negative classes \cite{rank2015}, i.e.:
\begin{equation}
c^- = \mathop{\argmax}_{c \in \mathcal{L}-L_k} \mathcal{F}(s,c)
\end{equation}

\noindent{$\bullet$ \textbf{with ATT (Variant-2)} \ \  Now we define the loss function for the option of \textbf{ATT} to combine sentences as follows:}
\begin{gather}
\nonumber G_{[att]} = \sum_{c^+ \in L_k} ( \rho [0,\sigma^+ - \mathcal{F}(s^{c^+},c^+)]_+ \\ + \rho [0,\sigma^- + \mathcal{F} (s^{c^+}, c^-)]_{+} )
\label{func11}
\end{gather}
where $s^{c}$ means the attention weights of representation $s$ are merged by comparing sentence embeddings with relation class $c$ and $c^-$ is chosen by the following function:
\begin{equation}
c^- = \mathop{\argmax}_{c \in \mathcal{L}-L_k} \mathcal{F}(s^{c^+},c)
\label{func13}
\end{equation}
which means we update one negative class in every training round.
We keep the values of $\rho$, $\sigma^+$ and $\sigma^-$ same as values in function (\ref{func9}).

According to this loss function, we can see that: for each class $c^+ \in L_k$, it will capture the most related information from sentences to merge $s^{c^+}$, then rank $\mathcal{F}(s^{c^+},c^+)$ higher than all negative scores which each is $\mathcal{F}(s^{c^+},c^-)$ ($c^- \in \mathcal{L} - L_k$). We use the same update algorithm to minimize this loss.

\noindent{$\bullet$ \textbf{Extended with ATT (Variant-3)} \ \ According to function (\ref{func11}), for each $c^+$, we only select one negative class to update the parameters, which only considers the connections between positive classes and negative ones, ignoring connections between positive classes, so we extend function (\ref{func11}) to better exploit class ties by considering the connections between positive classes. We give out the extended loss function as follows:}
\begin{gather}
\nonumber G_{[Exatt]} = \sum_{c^* \in L_k}( \sum_{c^+ \in L_k} \rho [0,\sigma^+ - \mathcal{F}(s^{c^*},c^+)]_+ \\ + \rho [0,\sigma^- + \mathcal{F} (s^{c^*}, c^-)]_{+} )
\label{func14}
\end{gather}
Similar to function (\ref{func13}), we select $c^-$ as follows:
\begin{equation}
c^- = \mathop{\argmax}_{c \in \mathcal{L}-L_k} \mathcal{F}(s^{c^*},c)
\label{func15}
\end{equation}
and we use the same method to update this loss function as discussed above. From the function (\ref{func14}), we can see that: for $c^* \in L_k$, after merging the bag representation $s$ with $c^*$, we share $s$ with all the other positive classes and update the class embeddings of other positive classes with $s$, in this way, the connections between positive classes can be captured and learned by our model. 

In loss function (\ref{func9}), (\ref{func11}) and (\ref{func14}), we combine losses from all positive labels to make joint extraction to capture the class ties among relations. 
Suppose we make separated extraction, the losses from positive labels will be divided apart and will not get enough information of connections between positive labels, comparing to joint extraction. 
Connections between positive labels and negative ones are exploited by controlling margins: $\sigma^+$ and $\sigma^-$.


\subsection{Relieving Impact of NR}
In relation extraction, the dataset will always contain certain negative samples which do not express relations classified as NR (not relation).
Table \ref{NR1} presents the proportion of NR samples in SemEval-2010 Task 8 dataset\footnote{This is a dataset for relation extraction in traditional supervision framework.} \cite{Semeval2010} and dataset from \citet{2010}, which shows almost data is about NR in the latter dataset. Data imbalance will severely affect the model training and cause the model only sensitive to classes with high proportion \cite{imbalance}.

\begin{table}{}
  \centering
  \begin{tabular}{|c|c|c|}
  \hline
  Pro. & Training & Test  \\ \hline
  SemE. & $17.63\%$ & $16.71\%$ \\ \hline
  Riedel & $72.52\%$ & $96.26\%$ \\ \hline
  \end{tabular}
  \caption{The proportions of NR samples from SemEval-2010 Task 8 dataset and Riedel dataset.}
  \label{NR1}
\end{table}

\IncMargin{0.em}
\begin{algorithm}
\SetKwData{Left}{left}\SetKwData{This}{this}\SetKwData{Up}{up}
\SetKwFunction{Union}{Union}\SetKwFunction{FindCompress}{FindCompress}
\SetKwInOut{Input}{input}\SetKwInOut{Output}{output}
\Input{$\mathcal{L}$, $(t_k, L_k, X_k)$ and $S_k$;}
\Output{$G_{[Exatt]}$;}
$G_{[Exatt]} \gets 0$\;
\For{$c^* \in L_k$}{\label{forins}
Merge representation $s^{c^*}$ by function (\ref{func5}), (\ref{func6}), (\ref{func7})\;

\For{$c^+ \in L_k$}{
\If{$c^+$ is not NR}{\label{ut}
$G_{[Exatt]} \gets G_{[Exatt]} + \rho[0, \sigma^+ - \mathcal{F}(s^{c^*},c^+)]_+$\;
}
}

$c^- \gets \argmax_{c \in \mathcal{L}-L_k} \mathcal{F}(s^{c^*}, c)$\;
$G_{[Exatt]} \gets G_{[Exatt]} + \rho [0, \sigma^- + \mathcal{F}(s^{c^*}, c^-)]_+$\;
}
\caption{Merging loss function of \textbf{Variant-3}}\label{algo-att}
return $G_{[Exatt]}$\;
\end{algorithm}\DecMargin{0.em}

In order to relieve the impact of NR in DS based relation extraction, we cut the propagation of loss from NR, which means if relation $c$ is NR, we set its loss as $0$. Our method is similar to \citet{rank2015} with slight variance. \citet{rank2015} directly omit the NR class embedding, but we keep it. If we use \textbf{ATT} method to combine information across sentences, we can not omit NR class embedding according to function (\ref{func6}) and (\ref{func7}), on the contrary, it will be updated from the negative classes' loss.

In Algorithm \ref{algo-att}, we give out the pseudocodes of merging loss with \textbf{Variant-3} and considering to relieve the impact of NR.


\section{Experiments}
\subsection{Dataset and Evaluation Criteria}
We conduct our experiments on a widely used dataset, developed by \citet{2010} and has been used by \citet{2011}, \citet{2012}, \citet{zeng2015} and \citet{lin2016}. The dataset aligns Freebase relation facts with the New York Times corpus, in which training mentions are from 2005-2006 corpus and test mentions from 2007.

Following \citet{2009}, we adopt held-out evaluation framework in all experiments. 
Aggregated precision/recall curves are drawn and precision@N (P@N) is reported to illustrate the model performance.
\subsection{Experimental Settings}
\noindent{\textbf{Word Embeddings.} \  \
We use a word2vec tool that is gensim\footnote{http://radimrehurek.com/gensim/models/word2vec.html} to train word embeddings on NYT corpus. Similar to \citet{lin2016}, we keep the words that appear more than $100$ times to construct word dictionary and use ``UNK'' to represent the other ones.}

\noindent{\textbf{Hyper-parameter Settings.} \ \
Three-fold validation on the training dataset is adopted to tune the parameters following \citet{2012}. We use grid search to determine the optimal hyper-parameters. We select word embedding size from $\{50,100,150,200,250,300\}$. Batch size is tuned from $\{80, 160, 320, 640\}$. We determine learning rate among $\{0.01,0.02,0.03,0.04\}$. The window size of convolution is tuned from $\{1, 3 ,5\}$. We keep other hyper-parameters same as \citet{zeng2015}: the number of kernels is $230$, position embedding size is $5$ and dropout rate is $0.5$. Table \ref{parameter} shows the detailed parameter settings.} 

\begin{table}
  \centering
  \begin{tabular}{|l|l|r|}
  \hline
  Parameter Name & Symbol & Value \\ \hline
  Window size & $d^{win}$& $3$ \\
  Sentence. emb. dim. & $d^f$ & $690$ \\
  Word. emb. dim. & $d^1$ & $50$ \\
  Position. emb. dim. & $d^2$ & $5$ \\
  Batch size & $\mathcal{B}$ & $160$ \\
  Learning rate & $\lambda$ & $0.03$ \\
  Dropout pos. & $p$ & $0.5$ \\ \hline
  \end{tabular}
  \caption{Hyper-parameter settings.}\label{parameter}
\end{table}

\subsection{Comparisons with Baselines}
\noindent{\textbf{Baseline.} \ \
We compare our model with the following baselines:}

$\bullet$ \textbf{Mintz} \cite{2009} \ the original distantly supervised model.

$\bullet$ \textbf{MultiR} \cite{2011} \ a multi-instance learning based graphical model which aims to address overlapping relation problem.

$\bullet$ \textbf{MIML} \cite{2012} \ also solving overlapping relations in a multi-instance multi-label framework.

$\bullet$ \textbf{PCNN+ATT} \cite{lin2016} \ the state-of-the-art model in dataset of \citet{2010} which applies \textbf{ATT} to combine the sentences.



\begin{figure}
\centering\includegraphics[width = 7.5cm]{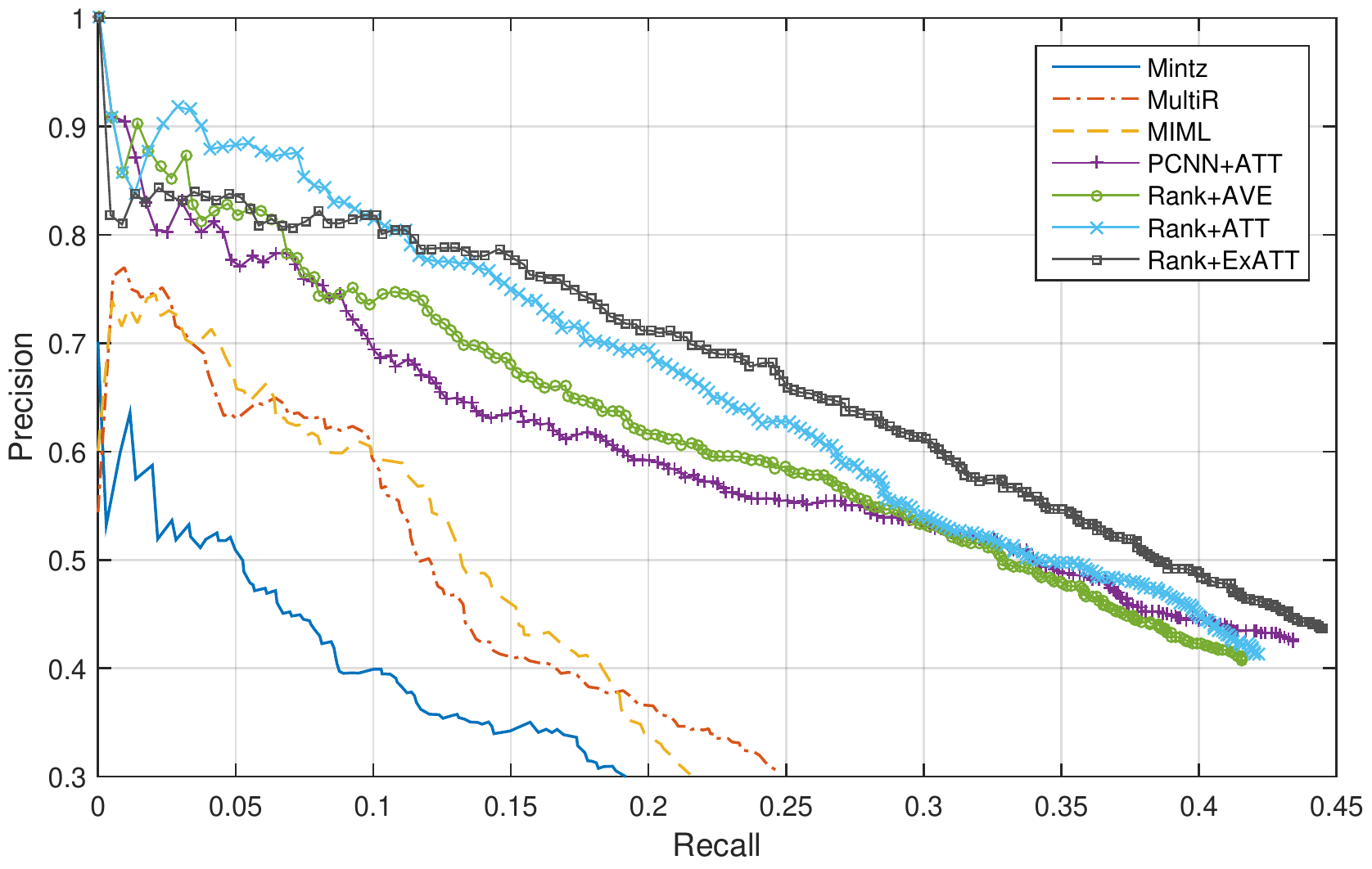}
\caption{\label{Fig:PC} Performance comparison of our model and the baselines.}
\end{figure}

\noindent{\textbf{Results and Discussion.} \ \
We compare our three variants of loss functions with the baselines and the results are shown in Figure \ref{Fig:PC}. From the results we can see that: (1) Rank + AVE (Variant-1) achieves comparable results with PCNN+ATT; (2) Rank + ATT (Variant-2) and Rank + ExATT (Variant-3) significantly outperform PCNN + ATT with much higher precision and slightly higher recall in whole view; (3) Rank + ExATT (Variant-3) exhibits the best performances comparing with all the other methods including PCNN + ATT, Rank + AVE and Rank + ATT.}

\subsection{Impact of Joint Extraction and Class Ties}
In this section, we conduct experiments to reveal the effectiveness of our model to learn class ties with three variant loss functions mentioned above, and the impact of class ties for relation extraction. As mentioned above, we make joint extraction to learn class ties, so to achieve the goal of this set of experiments, we compare joint extraction with separated extraction. To make separated extraction, we divide the labels of entity tuple into single label and for one relation label we only select the sentences expressing this relation,  then we use this dataset to train our model with the three variant loss functions. 
We conduct experiments with Rank + AVE (Variant-1), Rank + ATT (Variant-2) and Rank + ExATT (Variant-3) relieving impact of NR. Aggregated P/R curves are drawn and precisions@N ($100$, $200$, $\cdots$, $500$) are reported to show the model performances. 
\begin{figure}
\centering\includegraphics[width = 8.cm]{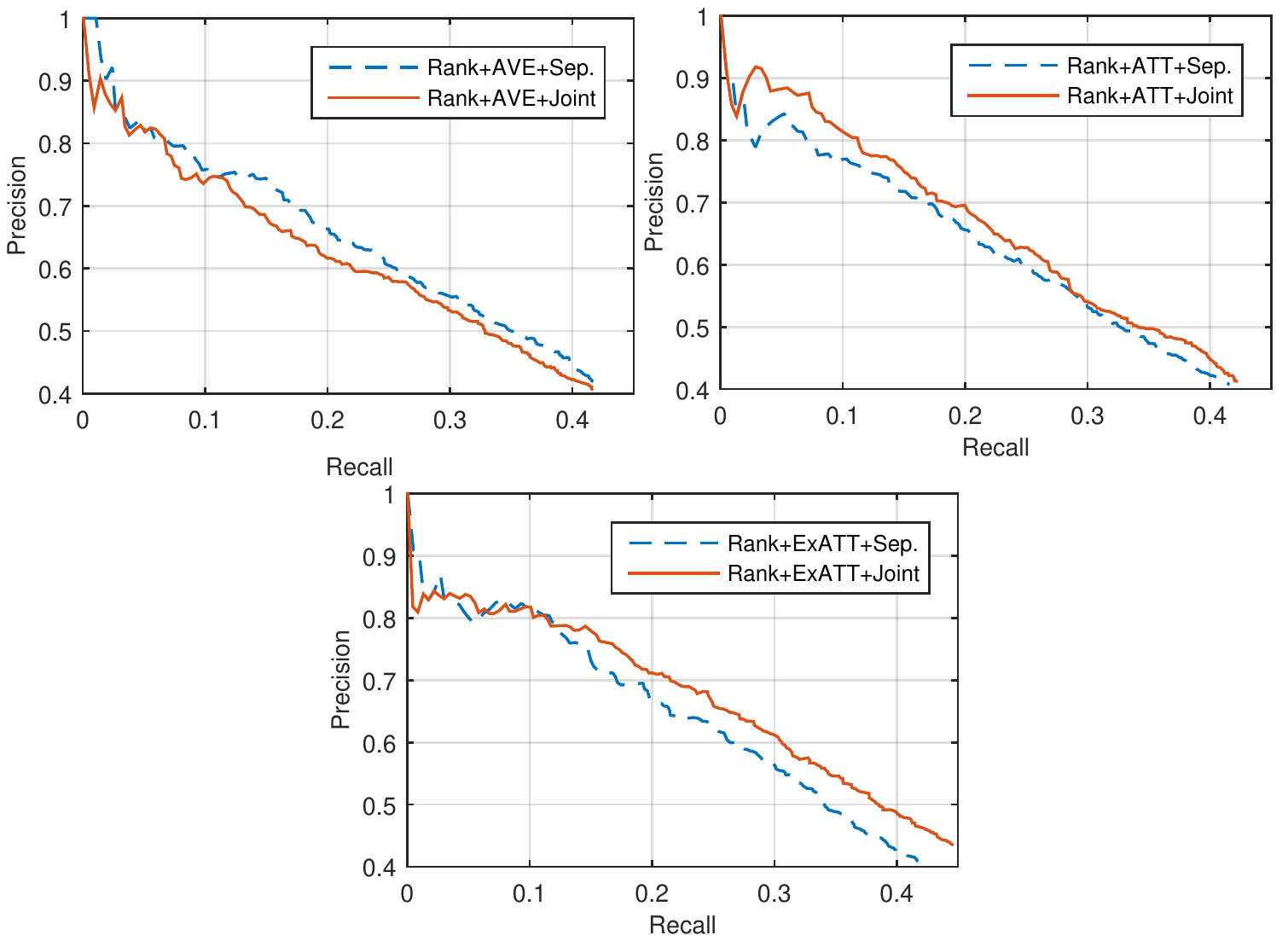}
\caption{Results for impact of joint extraction and class ties with methods of Rank + AVE, Rank + ATT and Rank + ExATT under the setting of relieving impact of NR.}
\label{JointEx}
\end{figure}

Experimental results are shown in Figure \ref{JointEx} and Table \ref{TabClassTies}. From the results we can see that: (1) For Rank + ATT and Rank + ExATT, joint extraction exhibits better performance than separated extraction, which demonstrates class ties will improve relation extraction and the two methods are effective to learn class ties; (2) For Rank + AVE, surprisingly joint extraction does not keep up with separated extraction. For the second phenomenon, the explanation may lie in the \textbf{AVE} method to aggregate sentences will incorporate noise data consistent with the finding in \citet{lin2016}. When make joint extraction, we will combine all sentences containing the same entity tuple no matter which class type is expressed, so it will engender much noise if we only combine them equally.



\begin{table}
  \centering
  \begin{tabular}{|p{2.05cm}|p{0.49cm}|p{0.49cm}|p{0.49cm}|p{0.49cm}|p{0.49cm}|p{0.5cm}|}
  \hline
  P@N(\%) & $100$ & $200$ & $300$ & $400$ & $500$ &Ave. \\ \hline
  \hline
  R.+AVE+J.  & $81.3$  & $76.4$	&  $74.6$	& $69.6$	& $66.0$ & $73.6$\\ 
  \hline
  R.+AVE+S. & \textbf{82.4} & \textbf{79.6}	& $74.6$	& \textbf{74.4}	& \textbf{69.9} & \textbf{76.2}\\ 
  \hline
  \hline
  R.+ATT+J. & \textbf{87.9} & \textbf{84.3} & \textbf{78.0} & \textbf{74.9}	& \textbf{70.3} & \textbf{79.1} \\ \hline
  R.+ATT+S. & $82.4$	& $79.1$	& $75.9$	& $71.9$	& $69.5$ & 75.7\\ \hline
  \hline
  R.+ExATT+J. & \textbf{83.5} & $82.2$ & $78.7$ & \textbf{77.2}	& \textbf{73.1} & \textbf{79.0} \\ \hline
  R.+ExATT+S. & $82.4$	& \textbf{82.7}	& \textbf{79.4}	& $74.2$	& $69.2$ & 77.6	 \\ \hline
  \end{tabular}
  \caption{Precisions for top $100$, $200$, $300$, $400$, $500$ and average of them for impact of joint extraction and class ties.}\label{TabClassTies}
\end{table}





\begin{figure}
\centering\includegraphics[width = 7.5cm]{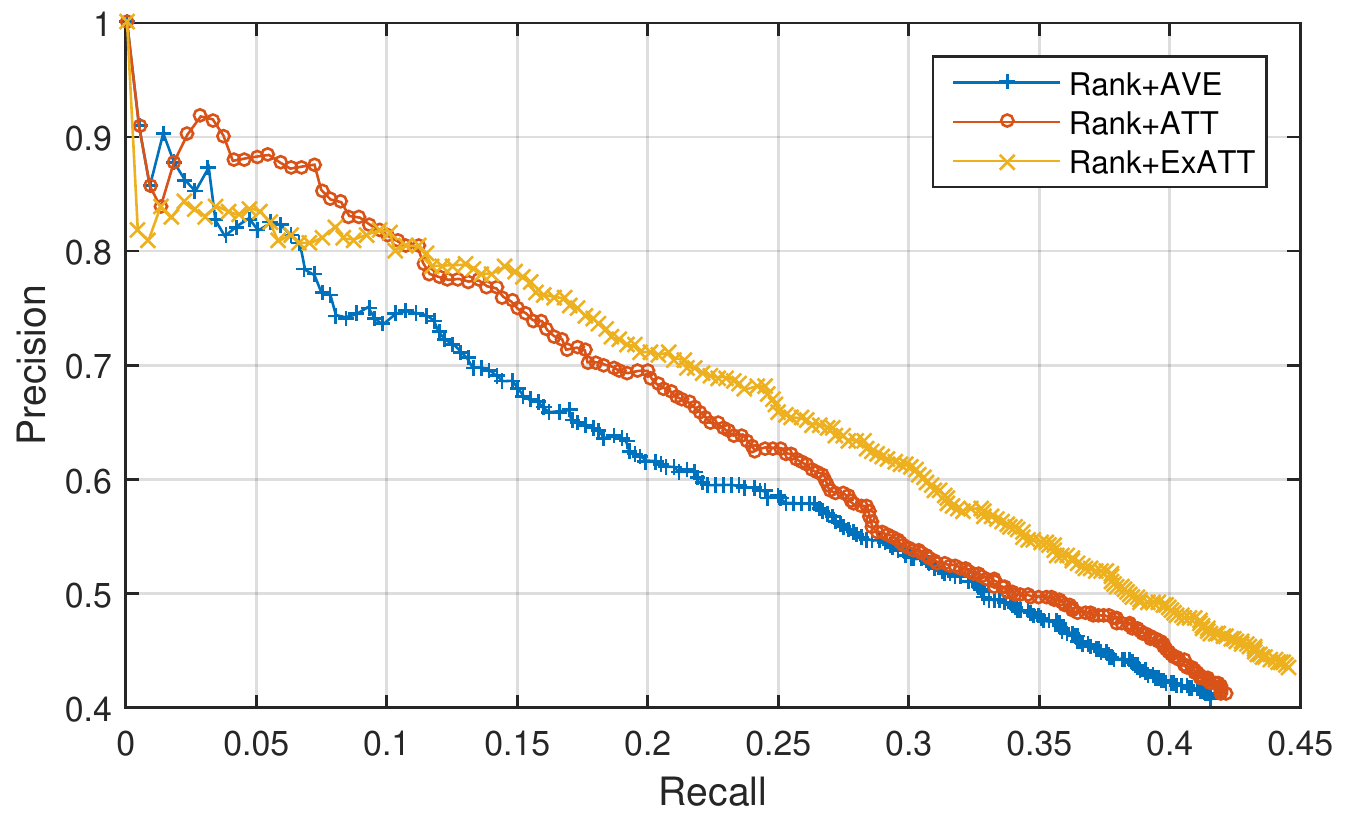}
\caption{Results for comparisons of variant joint extractions.}
\label{ATTAVE}
\end{figure}

\subsection{Comparisons of Variant Joint Extractions}
To make joint extraction, we have proposed three variant loss functions including Rank + AVE, Rank + ATT and Rank + ExATT in the above discussion and Figure \ref{Fig:PC} shows that the three variants achieve different performances. In this experiment, we aim to compare the three variants in detail. We conduct the experiments with the three variants under the setting of relieving impact of NR and joint extraction. We draw the P/R curves and report the top N ($100$, $200$, $\cdots$, $500$) precisions to compare model performance with the three variants. 

From the results as shown in Figure \ref{ATTAVE} and Table \ref{TabAVEATT} we can see that: (1) Comparing Rank + AVE with Rank + ATT, from the whole view, they can achieve the similar maximal recall point, but Rank + ATT exhibits higher precision in all range of recall; (2) Comparing Rank + ATT with Rank + ExATT, Rank + ExATT achieves much better performance with broader range of recall and higher precision in almost range of recall.



\begin{table}
  \centering
  \begin{tabular}{|p{1.6cm}|p{0.55cm}|p{0.55cm}|p{0.55cm}|p{0.55cm}|p{0.55cm}|p{0.55cm}|}
  \hline
  P@N(\%) & $100$ & $200$ & $300$ & $400$ & $500$ & Ave. \\ \hline
 \hline
  R.+AVE  & $81.3$  & $76.4$	&  $74.6$	& $69.6$	& $66.0$ & $73.6$ \\ 
  \hline
  R.+ATT & \textbf{87.9} & \textbf{84.3} & $78.0$ & $74.9$	& $70.3$ & \textbf{79.1} \\
  \hline
  R.+ExATT & $83.5$ & $82.2$ & \textbf{78.7} & \textbf{77.2}	& \textbf{73.1} & $79.0$\\ \hline
  \end{tabular}
  \caption{Precisions for top $100$, $200$, $300$, $400$, $500$ and average of them for Rank + AVE, Rank + ATT and Rank + ExATT.}\label{TabAVEATT}
\end{table}

\subsection{Impact of NR Relation}
\begin{figure}
\centering\includegraphics[width = 8.cm]{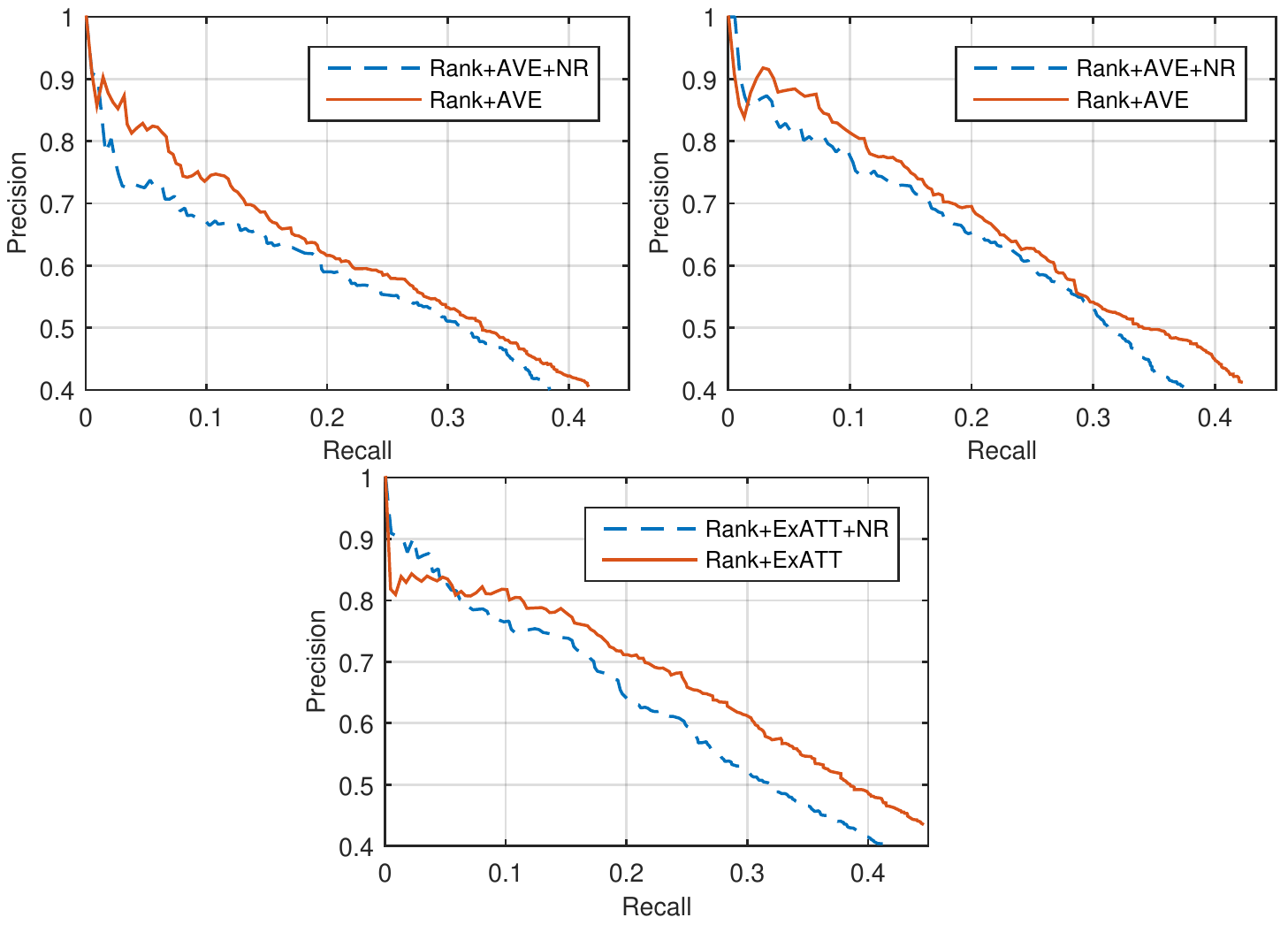}
\caption{Results for impact of relation NR with methods of Rank + AVE, Rank + ATT and Rank + ExATT.  ``+NR" means not relieving impact of NR.}
\label{NR}
\end{figure}

The goal of this experiment is to inspect how much relation of NR can affect the model performance. 
We use Rank + AVE, Rank + ATT, Rank + ExATT under the setting of relieving impact of NR or not to conduct experiments. We draw the aggregated P/R curves as shown in Figure \ref{NR}, from which we can see that after relieving the impact of NR, the model performance can be improved significantly. 

\begin{figure}
\centering\includegraphics[width = 8.cm]{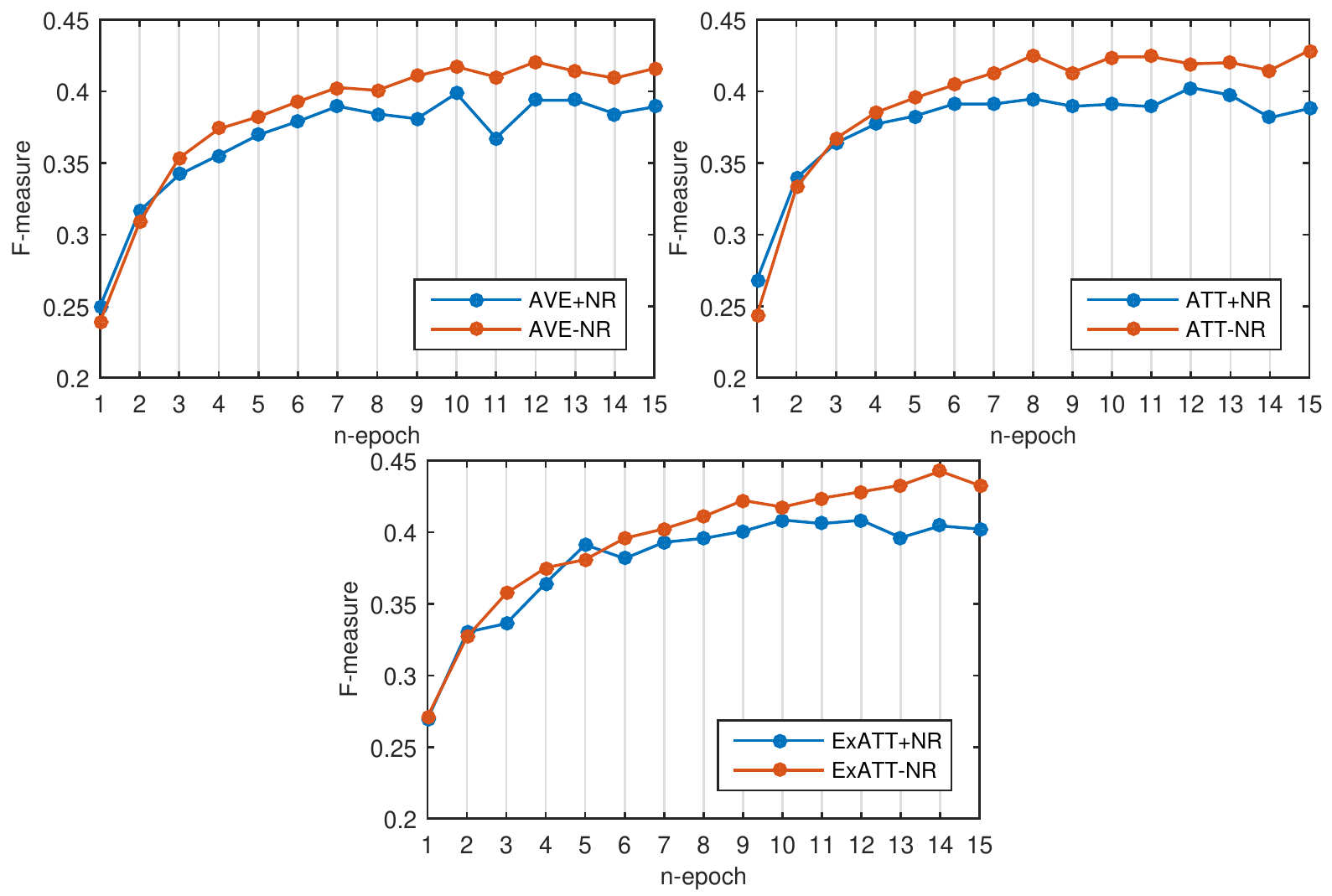}
\caption{\label{procedure} Impact of NR for model convergence. ``+NR" means not relieving NR impact; ``-NR" is opposite.}
\end{figure}

Then we further evaluate the impact of NR for convergence behavior of our model in model training. Also with the three variant loss functions, in each iteration, we record the maximal value of F-measure
\footnote{$F = 2 * P * R / (P + R)$}
to represent the model performance at current epoch. Model parameters are tuned for $15$ times and the convergence curves are shown in Figure \ref{procedure}. From the result, we can find out: ``+NR" converges quicker than ``-NR" and arrives to the final score at the around $11$ or $12$ epoch. In general, ``-NR" converges more smoothly and will achieve better performance than ``+NR" in the end.

\subsection{Case Study}
\begin{figure}
\centering\includegraphics[width = 7.5cm]{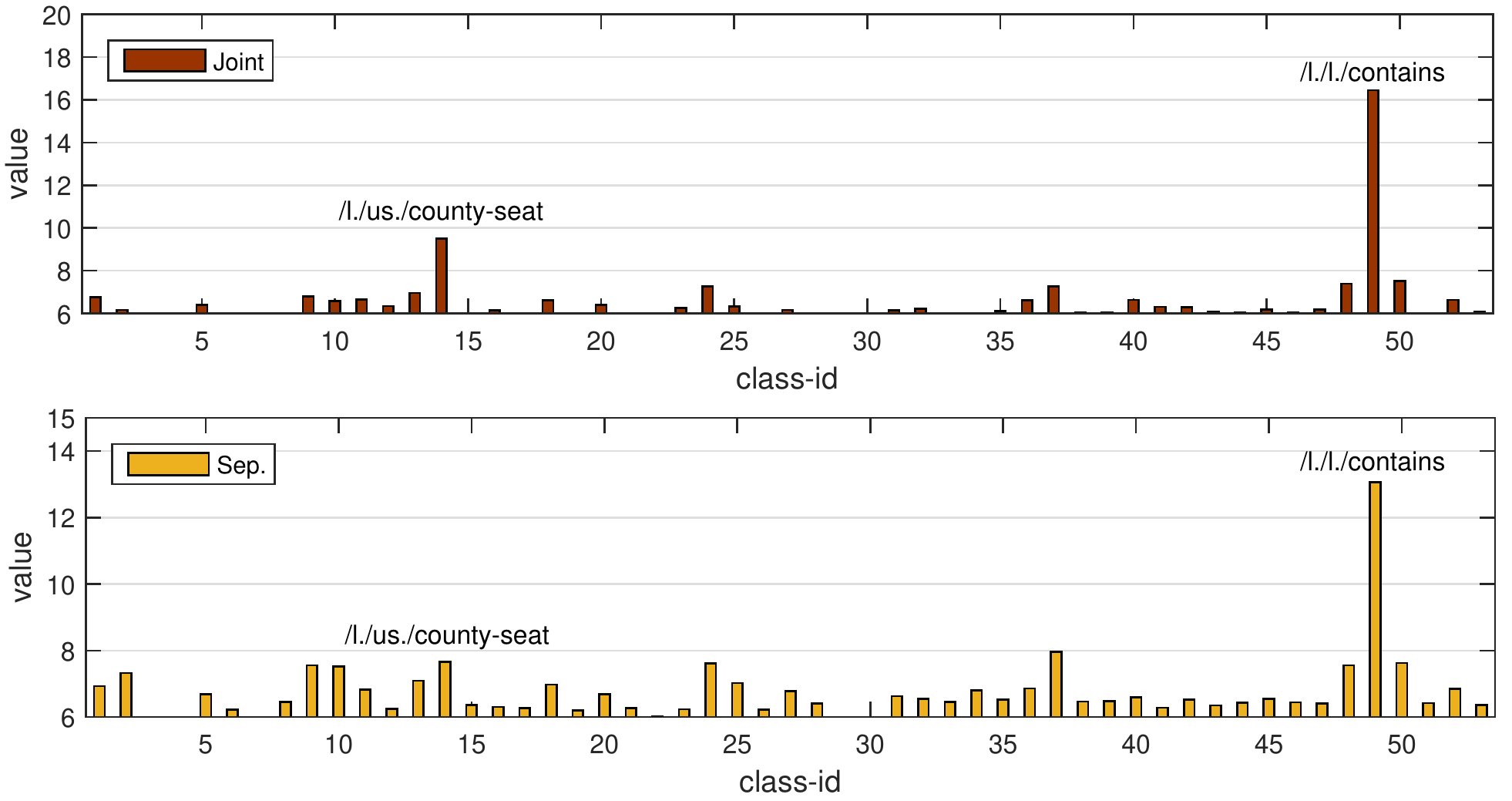}
\caption{\label{relation} The output scores for every relation with method of Rank + ATT. The top is under joint extraction setting; the bottom is under separated extraction.}
\end{figure}

\noindent{\textbf{Joint vs. Sep. Extraction (Class Ties).} \ \ We randomly select an entity tuple \emph{(Cuyahoga County, Cleveland)} from test set to see its scores for every relation class with the method of Rank + ATT under the setting of relieving impact of NR with joint extraction and separated extraction. This entity tuple have two relations: \emph{/location/./county\_seat} and \emph{/location/./contains}, which derive from the same root class and
they have weak class ties for
they all relating to topic of ``location''. We rescale the scores by adding value $10$. The results are shown in Figure \ref{relation}, from which we can see that: under joint extraction setting, the two gold relations have the highest scores among the other relations but under separated extraction setting, only \emph{/location/./contains} can be distinguished from the negative relations, which demonstrates that joint extraction is better than separated extraction by capturing the class ties between relations.}

\section{Conclusion and Future Works}
In this paper, we leverage class ties to enhance relation extraction by joint extraction using pairwise ranking combined with CNN. An effective method is proposed to relieve the impact of NR for model training. Experimental results on a widely used dataset show that leveraging class ties will enhance relation extraction and our model is effective to learn class ties. Our method significantly outperforms the baselines. 


In the future, we will focus on two aspects: (1) Our method in this paper considers pairwise intersections between labels, so to better exploit class ties, we will extend our method to exploit all other labels' influences on each relation for relation extraction, transferring \emph{second-order} to \emph{high-order} \cite{zhang2014review}; (2) We will focus on other problems by leveraging class ties between labels, specially on multi-label learning problems \cite{zhou2012multi} such as multi-category text categorization \cite{rousu2005learning} and multi-label image categorization \cite{zha2008joint}. 

\section*{Acknowledgments}
Firstly, we would like to thank Xianpei Han and Kang Liu for their valuable suggestions on the initial version of this paper, which have helped a lot to improve the paper. Secondly, we also want to express gratitudes to the anonymous reviewers for their hard work and kind comments, which will further improve our work in the future.
This work was supported by the National High-tech Research and Development Program (863 Program) (No. 2014AA015105) and National Natural Science Foundation of China (No. 61602490).


\bibliography{acl2017}
\bibliographystyle{acl_natbib}

\appendix

\end{document}